\documentclass[conference]{IEEEtran}

\IEEEoverridecommandlockouts                              % This command is only needed if 
\usepackage{helvet}         		% selects Helvetica as sans-serif font
\usepackage{type1cm}        		% activate if the above 3 fonts are not available on your system
\usepackage{graphics} 		% for pdf, bitmapped graphics files
\usepackage{graphicx}        % standard LaTeX graphics tool when including figure files
\usepackage[export]{adjustbox}
\usepackage{epsfig} 			% for postscript graphics files
\usepackage{multicol}        		% used for the two-column index
\usepackage[hang,flushmargin]{footmisc}
\usepackage{amsmath}
\usepackage{amssymb}
\usepackage[ruled]{algorithm2e}
\usepackage{bm}
\usepackage{newtxtext}       % 
\usepackage{newtxmath}       % selects Times Roman as basic font
\usepackage{caption}
\usepackage{gensymb}
\usepackage{booktabs}
\usepackage{epstopdf}
\usepackage{url}
\usepackage[font={small}]{caption}
\usepackage{commath}
\usepackage{subfigure}
\usepackage{multirow}
\usepackage{makecell}
\usepackage{calc}
\usepackage{pifont} 
\usepackage{tikz}

\title{\LARGE \bf \vspace{5mm}
DRACo-SLAM2: Distributed Robust Acoustic Communication-efficient SLAM for Imaging Sonar Equipped Underwater Robot Teams with Object Graph Matching
\vspace{-4mm}}

\author{
Yewei Huang$^{1*}$, John McConnell$^{2*}$, Xi Lin$^{1}$ and Brendan Englot$^{1}$% <-this % stops a space
\thanks{* Equal contribution.}
\thanks{$^{1}$Y. Huang, X. Lin and B. Englot are with Stevens Institute of Technology, Hoboken, NJ, USA, {\tt\footnotesize \{yhuang85, xlin26, benglot\}@stevens.edu}. $^{2}$J. McConnell is with the U.S. Naval Academy, Annapolis, MD, USA, {\tt\footnotesize jmcconne@usna.edu}. This research was supported in part by NSF Grant IIS-1652064 and ONR Grant N00014-24-1-2522.}
}
\begin{document}
\maketitle
\thispagestyle{empty}
\pagestyle{empty}
\newcommand{\edit}[1]{\textcolor{blue}{#1}}

%%%%%%%%%%%%%%%%%%%%%%%%%%%%%%%%%%%%%%%%%%%%%%%%%%%%%%%%%%%%%%%%%%%%%%%%%%%%%%%%
\begin{abstract}
We present DRACo-SLAM2, a distributed SLAM framework for underwater robot teams equipped with multibeam imaging sonar.
This framework improves upon the original DRACo-SLAM by introducing a novel representation of sonar maps as object graphs and utilizing object graph matching to achieve time-efficient inter-robot loop closure detection without relying on prior geometric information. 
To better-accommodate the needs and characteristics of underwater scan matching, we propose incremental \textit{Group-wise} Consistent Measurement Set Maximization (GCM), a modification of Pairwise Consistent Measurement Set Maximization (PCM), which effectively handles scenarios where nearby inter-robot loop closures share similar registration errors. The proposed approach is validated through extensive comparative analyses on simulated and real-world datasets.
\end{abstract}

%%%%%%%%%%%%%%%%%%%%%%%%%%%%%%%%%%%%%%%%%%%%%%%%%%%%%%%%%%%%%%%%%%%%%%%%%%%%%%%%

\section{Introduction}
Exploring the %vast and mysterious 
ocean has long been one of humanity's greatest ambitions, and robust and efficient simultaneous localization and mapping (SLAM) algorithms can provide a cornerstone for achieving this efficiently with large teams of robots. Multi-robot SLAM is also valuable for executing other long-term and large-scale marine missions, such as constructing and maintaining offshore energy infrastructure, monitoring water quality over extended periods \cite{wang2022cooperation}, and conducting targeted sampling near the ocean floor~\cite{zhang2024coordinated}.

Although a variety of single-robot SLAM algorithms utilizing sonar \cite{wang2017underwater, mcconnell2024largescale}, cameras \cite{song2024turtlmap}, or their combination \cite{rahman2022svin2} have been developed, there are still significant challenges facing multi-robot SLAM in underwater environments. Chief among these is the limited bandwidth available for wireless communication. For example, the HS underwater acoustic modems\footnote{https://www.evologics.com/acoustic-modem/hs}, commonly used for subsea communication, provide only 62.5 kbps even for short-range transmissions. 
Therefore, data efficiency is essential for underwater multi-robot SLAM.

%Unlike most other robotics applications, in underwater environments, communication bandwidth poses a significant challenge for multi-robot SLAM. 

Although multi-robot SLAM has been applied in underwater environments using cameras \cite{bonin2020towards, xanthidis2022towards}, the effectiveness of cameras is limited to clear water conditions \cite{wang2019underwater}, which are uncommon in sandy, near-coastal areas or regions with ongoing construction activities.
Thus, a need remains for sonar-based robot teams in many real-world applications. 
Unlike cameras, which provide clear, high-resolution images, sonar images are affected by acoustic noise and have limited resolution. Registering sonar images without prior geometric information is consequently more difficult. Therefore, effective inter-robot loop closure detection remains a key challenge in underwater sonar-based SLAM systems.

\begin{figure}
        \centering
        \includegraphics[width=.98\columnwidth]{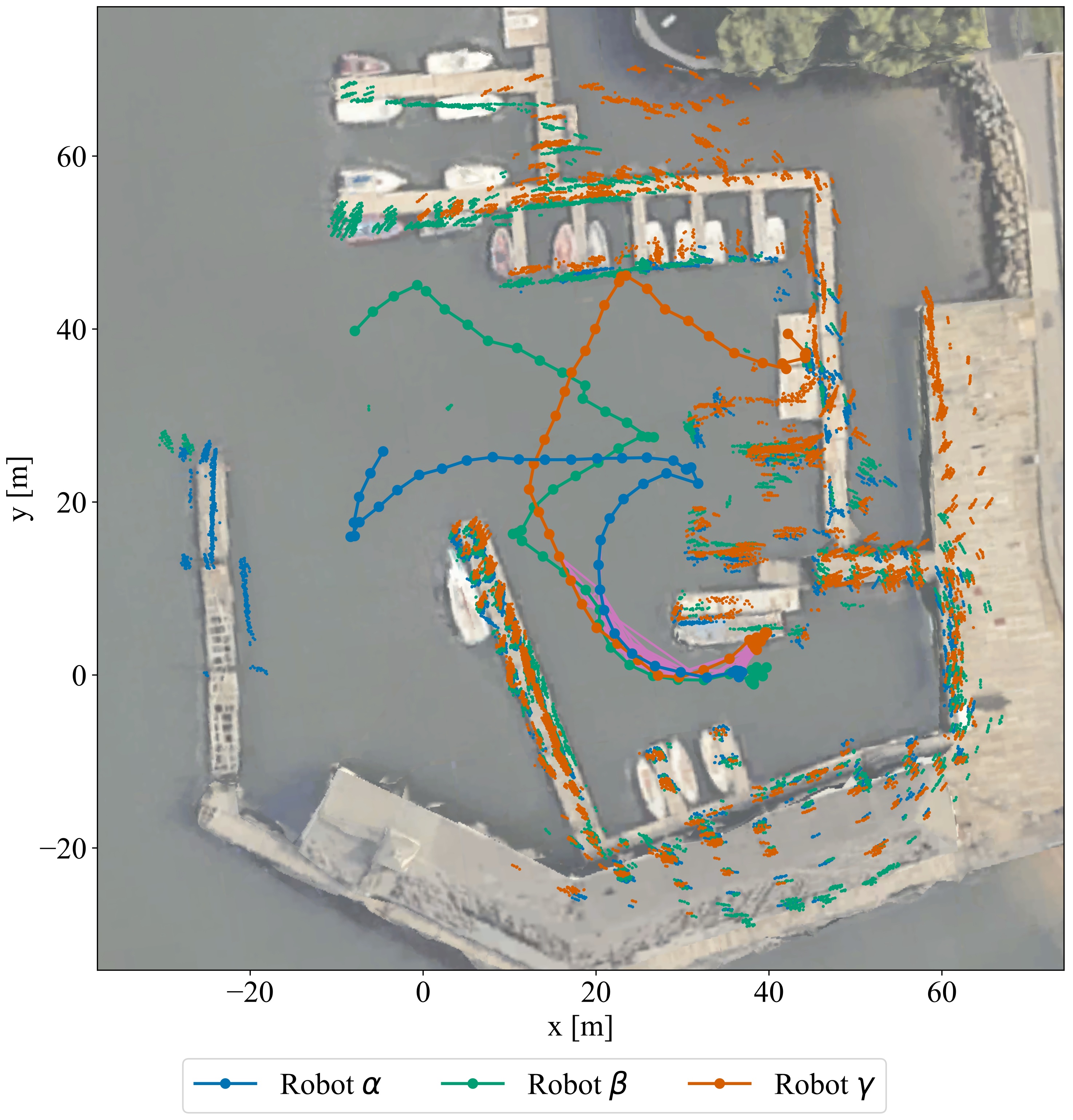}
        \caption{\textbf{Example DRACo-SLAM2 result with real sonar data.} Optimized trajectories and point clouds from three robots using the proposed DRACo-SLAM2 on a dataset collected at the U.S. Merchant Marine Academy, King’s Point, NY, aligned with a satellite image. Inter-robot measurement constraints are shown in purple.}\label{fig:real}
        \label{fig:usmma_real}
        \vspace{-5mm}
\end{figure}
We demonstrated the feasibility of performing inter-robot loop closure detection and multi-robot SLAM using only %acoustic 
sonar images in DRACo-SLAM \cite{mcconnell2022draco}. 
However, the global Iterative Closest Point (Go-ICP) algorithm \cite{yang2013goicp} used in that system is computationally expensive, limiting the overall system to operate at relatively low frequencies.
In this paper, we introduce DRACo-SLAM2, an enhanced multi-robot SLAM system for underwater environments that achieves greater computational efficiency through object graph matching.

DRACo-SLAM2 constructs an object map of the surrounding environment based on pose estimation from local sonar SLAM, which is then shared among robot team members for object graph matching.
Relative transformations between object maps serve as initial guesses for ICP registration, eliminating the need for Go-ICP in this approach and significantly reducing computation time. 
To detect outliers among potential inter-robot loop closures, we propose and employ \textit{Group-wise} Consistent Measurement Set Maximization (GCM) in lieu of PCM \cite{mangelson2018pairwise}. 
%Once inter-robot loop closures are verified, a two-step local and global optimization process is performed to ensure stable state estimation.

Our contributions are summarized as follows:  
\begin{itemize}  
    \item We present the first application of object graph matching for inter-robot loop closure detection using sonar images in underwater environments.
    \item The proposed object graph matching provides a reliable initial guess for refined ICP scan registration, enabling the detection of more inter-robot loop closures in a computationally efficient manner.  
    \item We introduce the Group-wise Consistent Measurement Set Maximization (GCM) algorithm for robust loop closure selection, targeting cases where multiple loop closures exhibit similar registration errors.  
\end{itemize}  

Our code for the proposed framework and the datasets described in this paper {is publicly released\footnote{\url{https://github.com/RobustFieldAutonomyLab/DRACO-SLAM2}}}.  
The subsequent sections of this paper are organized as follows: 
Sec.~\ref{sec:related}~provides a review of the background literature,  
followed by a detailed description of our proposed multi-robot SLAM algorithm in Sec.~\ref{sec:Algorithm}. 
Experimental results are presented in Sec.~\ref{sec:setup}, and the paper concludes with a summary in Sec.~\ref{sec:conclusion}.

\begin{figure*}[htp]
    \centering
    \includegraphics[width=.9\textwidth]{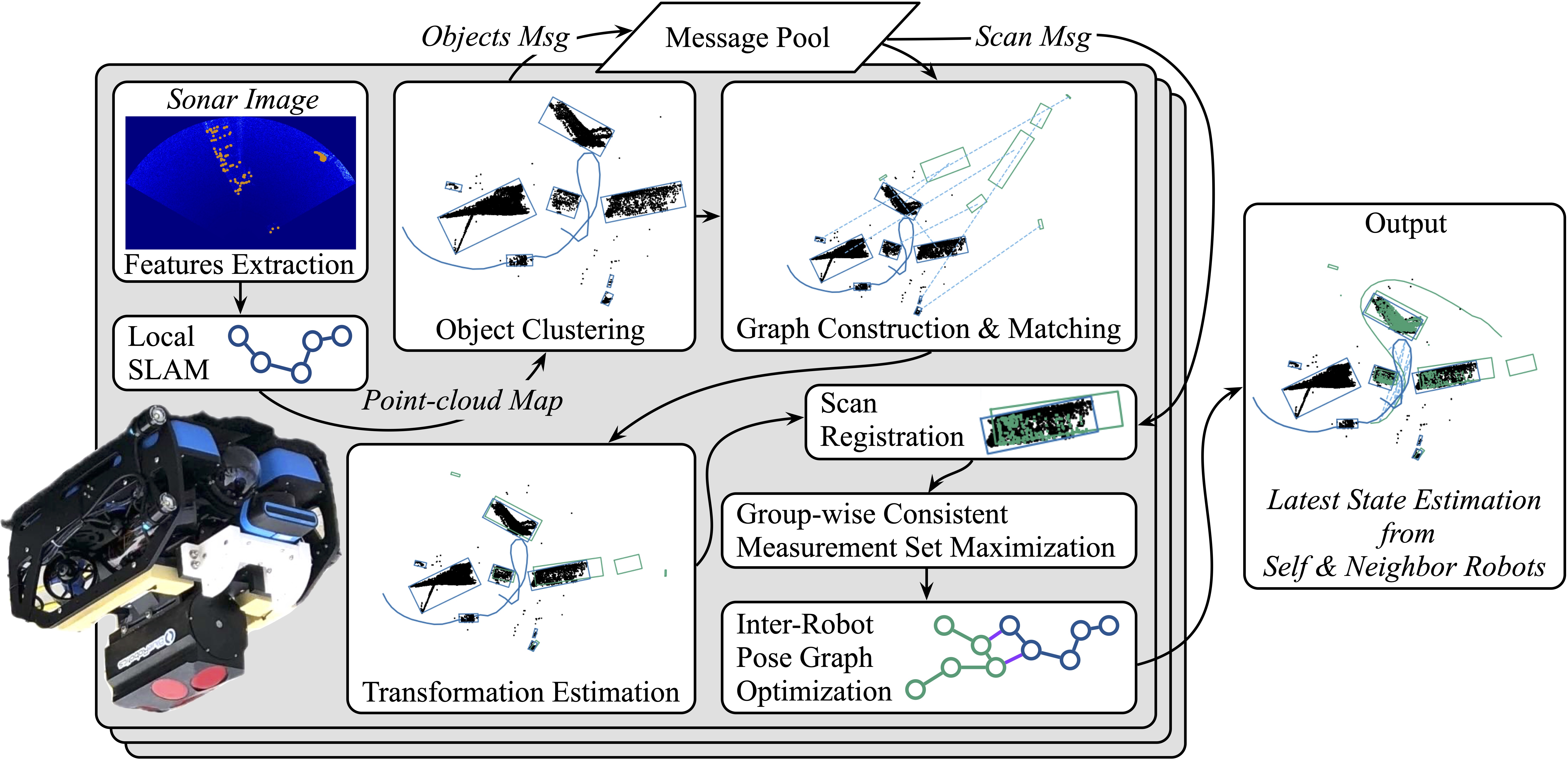}
    \caption{\textbf{Overview of DRACo-SLAM2 Architecture.
    %Proposed inter-robot SLAM with object graph matching on the plane dataset
    } Sonar images are captured by a horizontally-oriented sonar. Each robot's object map is clustered using DBSCAN. The local robot receives the neighboring robot's object map, aligns it to its local map using graph matching, and requests scans for ICP registration with the graph matching transformation as an initial guess. Inter-robot loop closures are then added to the pose graph for the two-step pose graph optimization (PGO). The local robot's trajectory and object vertices are shown in blue, while the neighboring robot's data is in green. Dashed lines indicate detected correspondences between objects in different maps.}
    \label{fig:pipeline}
    \vspace{-5mm}
\end{figure*}
\section{Related Works}\label{sec:related}
One key challenge that makes the underwater multi-robot SLAM problem unique, compared to other SLAM problems, is the limited communication bandwidth resulting from the marine environment.
%This constraint has led many researchers to address the underwater multi-robot SLAM problem using various approaches.
To address this constraint, Bonin-Font et al.~\cite{bonin2020towards} and $\text{MAM}^3$SLAM \cite{drupt2024mam3slam} propose centralized algorithms, where a server handles all inter-robot data association and optimization under the assumption of ``a powerful server without communication restrictions". 
However, these methods are only tested on small-range datasets due to the communication bandwidth degradation as robots move farther from the server.

Zhang et al.~\cite{zhang2023underwater} avoids the communication problem by employing a distributed local SLAM data collection approach combined with a GMRBnB-based map registration strategy, similar to the multi-session SLAM frameworks~\cite{bryson2013automated, williams2016return}. 
While these methods are effective for bathymetric mapping tasks, they are less suitable for real-time decision-making applications, such as active planning and exploration.

Paull et al.~\cite{paull2015communication} consider a scenario where robots observe one another and share their landmark observations. In their approach, both landmark and robot states are optimized in a distributed manner across team members using a graph-based representation.
Özkahraman et al.~\cite{ozkahraman2022collaborative} explore a similar situation, intentionally designing robot rendezvous strategies to achieve more accurate localization in environments where landmarks are absent.
However, achieving robot rendezvous is challenging, particularly for long-term missions, due to factors such as the vastness of marine environments, unpredictable oceanic currents, and time synchronization issues.

In this paper, we adopt the communication assumptions described in DRACo-SLAM~\cite{mcconnell2022draco}, where robots can communicate with each other (at low bandwidth, using wireless acoustic communication) when within a relatively long distance range. This assumption enables the real-time estimation of both self and neighbor robot states across a relatively large environment.

Another key consideration is the inter-robot data association or map merging strategy for multi-robot teams. 
Most approaches rely on extracting visual features from camera or sonar images. 
Bryson et al. \cite{bryson2013automated} extract SIFT features from camera images and perform inter-robot structure-from-motion (SfM). 
Bonin-Font et al. \cite{bonin2020towards} also extract SIFT features from camera images but utilize HALOC as the feature descriptor.
$\text{MAM}^3$SLAM~\cite{drupt2024mam3slam} extracts ORB features and associates keyframes using DBoW2. 
Zhang et al. \cite{zhang2023underwater} and Gaspar et al. \cite{gaspar2023feature} extract ORB features from sonar images and adopt a similar visual vocabulary strategy. 
While these feature-based methods are straightforward, their high-dimensional feature descriptors place significant demands on communication bandwidth.

To alleviate the communication bandwidth burden caused by high-dimensional feature descriptors, some methods use compact features to represent entire visual or sonar images. 
Paull et al. \cite{paull2015communication} extract mine-like objects from sonar images and exchange their positions among team members.
Santos et al. \cite{santos2019underwater} extract objects from sonar images and utilize scene graphs to describe and match sonar data. 
DRACo-SLAM~\cite{mcconnell2022draco} and Sonar-context~\cite{kim2023robust} represent sonar images with compact sonar descriptors for efficient data exchange and loop closure candidate detection. 
While compact features significantly reduce the bandwidth required for inter-robot data exchange, they can lead to errors %mis-alignments 
in regions with repeating patterns (i.e., perceptual aliasing).

Some methods perform data association using features extracted from a larger %an entire local 
submap to further mitigate perceptual aliasing. %the local optima caused by repeating patterns. 
Wang et al. \cite{wang2022cooperation} apply the DBScan algorithm to cluster and describe the shapes of oil spills in the ocean.
Deng et al. \cite{deng2024underwater} utilize a deep learning-based encoder-decoder structure for point cloud transmission.
Qi et al.~\cite{qi2024efficient} employ a terrain description feature to characterize submaps and match them using a genetic algorithm (GA). However, the GA process is computationally expensive and challenging to tune.

Other methods focus on robust loop closure outlier detection algorithms. 
Most approaches use a RANSAC strategy for map merging. 
DRACo-SLAM~\cite{mcconnell2022draco} uses pairwise consistent measurement set maximization (PCM) \cite{mangelson2018pairwise} to reduce %random 
outliers introduced by repeating patterns. 
However, despite its emphasis of measurement consistency, PCM is not robust to outliers characterized by high data similarity. %due to perceptual aliasing. %that follow similar registration errors. 
To address this issue, Do et al. \cite{do2020robust} propose a probabilistic approach specifically tailored for underwater scenarios to robustly reject outliers by seeking both consistency and similarity.  %that result from similar registration errors.

Inspired by these methods, we propose DRACo-SLAM2, a multi-robot SLAM framework that uses object graph matching. 
We cluster objects from the local SLAM map using the DBSCAN clustering algorithm and initially exchange \textit{only} the object map with team members to reduce computational burden and avoid misaligned loop closures caused by repeating patterns. Additionally, a group-wise consistent measurement set maximization (GCM) is introduced to enhance outlier detection in inter-robot data association, by fully utilizing the information obtained from the object graph matching.

\section{DRACo-SLAM2 Algorithm}\label{sec:Algorithm} 
In this section, we provide a comprehensive introduction to DRACo-SLAM2, the proposed multi-robot %inter-robot 
SLAM framework utilizing object graph matching.
Fig. \ref{fig:pipeline} presents the complete pipeline of the proposed framework. %demonstrated using the plane dataset as an example. The plane dataset was generated using the UUV simulator \cite{manhaes2016uuv} within the Gazebo simulation environment \cite{koenig2004design}. A detailed description of our simulated dataset is provided in Sec.~\ref{sec:setup}.
%\textbf{Overview.} 
We define the local robot as robot $\alpha$. 
The inter-robot data association process begins by clustering the latest point-cloud map, $\mathcal{M}{\alpha}$, generated by the local SLAM algorithm into an object map, $\mathcal{V}{\alpha}$. 
The object map is then shared among the robot team.

When the local robot receives an object map, $\mathcal{V}{\beta}$, from a neighboring robot $\beta$, the object maps are matched through object graph matching.
A detailed description of the object graph matching process is provided in Sec.~\ref{sec:graph}.
Feature points from the scans corresponding to overlapping objects are then requested from robot $\beta$ for performing scan registration.
The transformation $\mathbf{T}^{\beta}_{\alpha}$, representing the coordinate transformation from $\mathcal{V}{\beta}$ to $\mathcal{V}{\alpha}$, serves as the initial estimate for scan registration, as detailed in Sec.~\ref{sec:registration}.

After registration, incremental group-wise consistent measurement set maximization (GCM) is applied to ensure robustness by mitigating errors caused by %similar registration inaccuracies from adjacent sonar scans. 
perceptual aliasing.
All loop closures that pass the GCM process are subsequently added to the factor graph. A two-step inter-robot pose graph optimization is then performed to achieve a stable result.
The detailed process is presented in Sec.~\ref{sec:inter-PGO}.

\subsection{Local SLAM, Point-cloud Map and Object Map}
\subsubsection{Local SLAM}
We choose Bruce-SLAM \cite{wang2022virtual} as our local SLAM framework.
Bruce-SLAM is a sonar SLAM algorithm that uses data exclusively from forward-looking sonar (FLS) and vehicle dead-reckoning measurements.
However, the flexibility of our proposed DRACo-SLAM2 system allows for seamless integration with any %\textcolor{red}{2D or 3D} 
local sonar SLAM algorithm.

As a graph-based SLAM algorithm, Bruce SLAM solves the SLAM optimization problem as a maximum a posteriori (MAP) estimation problem \cite{isam22011Kaess}:
\begin{align}
\mathcal{X}^*_{\alpha} &= \arg \max_{\mathcal{X}_{\alpha}} P(\mathcal{X}_{\alpha} \mid \mathcal{Z}_{\alpha}),\label{eqn:local}\\
\mathcal{Z}_{\alpha} &= \mathcal{Z}^o_{\alpha} \cup \mathcal{Z}^s_{\alpha} \cup \mathcal{Z}^l_{\alpha},\label{eqn:z_local}
\end{align}
where $\mathcal{X}_{\alpha}$ represents the states of robot $\alpha$'s entire history, and $\mathcal{Z}_{\alpha}$ denotes the set of observations, including odometry measurements from the vehicle's dead reckoning system ($\mathcal{Z}^o_{\alpha}$), sequential scan matching measurements from neighboring sonar images ($\mathcal{Z}^s_{\alpha}$), and intra-robot loop closure measurements ($\mathcal{Z}^l_{\alpha}$).  

\subsubsection{Point-cloud map} To perform sequential scan matching and intra-robot data association, feature points are extracted from sonar images using SOCA-CFAR \cite{El-Darymli2013}, a variant of the Constant False Alarm Rate (CFAR) technique \cite{richards2005fundamentals}. 
Let the set of feature points detected from the sonar image captured at timestamp $i$ by robot $\alpha$ be denoted as $\mathcal{F}_{\alpha_i}$.
These points are transformed from the sensor frame into the local frame of robot $\alpha$, with state $\mathbf{x}_{\alpha_i} \in \mathcal{X}_{\alpha}$, and are represented as $_\alpha \mathcal{F}_{\alpha_i}$.
A local point-cloud map for robot $\alpha$, shown as black points in Fig. \ref{fig:pipeline}, is thus defined as $\mathcal{M}_{\alpha} = \{_\alpha \mathcal{F}_{\alpha_i} \mid i \in [0,t]\}$, where $t$ represents the current timestamp.

\subsubsection{Object map} 
The DBSCAN clustering algorithm \cite{ester1996density} is applied to cluster objects from the latest local point-cloud map, $\mathcal{M}_{\alpha}$.
For each cluster, a bounding rectangle is computed based on the positions of the points in the cluster.
The bounding rectangles of local objects are shown in blue in Fig. \ref{fig:pipeline}.
Each object, denoted as $\mathbf{v}_{\alpha_i}$, is described by the center coordinates $(x_{\alpha_i}, y_{\alpha_i})$ and the dimensions (length and breadth) $(l_{\alpha_i}, b_{\alpha_i})$ of its bounding rectangle.
To filter out noise misidentified as objects, only clusters with a number of points $n_p > n_{\text{min}}$ and a dimension $\max(l_{\alpha_i}, b_{\alpha_i}) > d_{\text{min}}$ are accepted.
The set of all accepted objects forms the object map, $\mathcal{V}_{\alpha}$.
 
\subsection{Object Graph Construction and Matching}\label{sec:graph}
When the object map $\mathcal{V}_{\beta}$ is received from the neighboring robot $\beta$, object graph $\mathcal{G}_{\beta} = (\mathcal{V}_{\beta}, \mathcal{E}_{\beta})$ is constructed using $\mathcal{V}_{\beta}$ as vertices in the graph. 
$\mathcal{G}_{\beta}$ is a directed complete graph, and $\mathcal{E}_{\beta}$ is further defined as:  
\begin{equation}
    \mathcal{E}_{\beta} = \{\mathbf{e}_{ij} = (\mathbf{v}_i, \mathbf{v}_j, w_{ij}) \mid \mathbf{v}_i, \mathbf{v}_j \in \mathcal{V}_{\beta}\},
\end{equation}
where $w_{ij}$ is the Euclidean distance between the centers of objects $\mathbf{v}_i$ and $\mathbf{v}_j$.
A local object graph $\mathcal{G}_{\alpha}$ is constructed in the same manner.
Thus, the object map matching problem is converted into a bipartite graph matching problem between $\mathcal{G}_{\alpha}$ and $\mathcal{G}_{\beta}$.

Inspired by SemanticLoop \cite{yu2022semanticloop}, we formulate this bipartite graph matching problem as a Quadratic Assignment Problem (QAP):
\begin{align}
    \mathbf{A}_v^* &= \arg \max_{\mathbf{A}_v} \sum_{\mathbf{e}_{ij} \in \mathcal{E}_\alpha}\sum_{\mathbf{e}_{kl} \in \mathcal{E}_\beta} a_{ik} \cdot a_{jl} \cdot u(\mathbf{e}_{ij}, \mathbf{e}_{kl}),\label{eqn:QAP}\\
    & = \arg \max_{\mathbf{A}_v} \left[\mathrm{vec}(\mathbf{A}_v)^\top \mathbf{U}_e \mathrm{vec}(\mathbf{A}_v)\right],\label{eqn:linear}\\
    u(\mathbf{e}_{ij}, \mathbf{e}_{kl}) & = \exp \left( - \mu |w_{ij} - w_{kl}| - |l_{i} - l_{k}| - |b_{i} - b_{k}| \right).
\end{align}
Let $\mathbf{A}_v$ be the boolean vertex assignment matrix with size $n_\alpha \times n_\beta$, where $n_\alpha$ and $n_\beta$ denote the numbers of vertices in $\mathcal{G}_{\alpha}$ and $\mathcal{G}_{\beta}$, respectively. 
Eq.~\eqref{eqn:linear} provides a vectorized reformulation of Eq.~\eqref{eqn:QAP}, where 
$\mathrm{vec}(\mathbf{A}_v)$ represents the vectorization of $\mathbf{A}_v$.
The utility matrix $\mathbf{U}_e$ contains the utility function $u(\cdot)$ as its elements, evaluating the similarity between edge pairs.

One key difference between our method and SemanticLoop \cite{yu2022semanticloop} is the calculation of the utility function $u(\cdot)$.
While SemanticLoop considers the semantic classes of objects and the Euclidean distances between their centers, our approach compares object dimensions and corresponding edge weights, with $\mu = 4$ serving as a scaling factor.
We add further constraints to ensure that each vertex in $\mathcal{G}_{\alpha}$ matches with at most one vertex in $\mathcal{G}_{\beta}$:
\begin{equation}
\forall i \leq n_\alpha, \sum_{j = 1}^{n_\beta} a_{ij} \leq 1, 
\forall j \leq n_\beta, \sum_{i = 1}^{n_\alpha} a_{ij} \leq 1. 
\end{equation}

We relax the boolean constraints on $ \mathbf{A}_v $, such that $ 0 \leq a_{ij} \leq 1 $, yielding $ \mathbf{A}_e $, which represents a relaxed version of $ \mathbf{A}_v $. This relaxation allows Eq.~\eqref{eqn:QAP} to be solved using the spectral method \cite{richards2005fundamentals}:
\begin{equation}
    \mathbf{U}_e \, \mathrm{vec}(\mathbf{A}_e) = \lambda_{\max} \, \mathrm{vec}(\mathbf{A}_e).
\end{equation}
The optimized solution, $ \mathrm{vec}(\mathbf{A}_e) $, is the eigenvector corresponding to the largest eigenvalue $ \lambda_{\max} $ of $ \mathbf{U}_e $. 

Next, $ \mathbf{A}_e $, which serves as a vertex similarity weight matrix, can be used to compute $ \mathbf{A}_v $ as follows:
\begin{align}\label{eq:lap}
    \mathbf{A}_v^* = \arg \min_{\mathbf{A}_v} \sum_{i = 1}^{n_\alpha} \sum_{j = 1}^{n_\beta} (-c_{ij} \cdot a_{ij}),
\end{align}
where $ c_{ij} $ denotes the respective element from $ \mathbf{A}_e $. We solve this Linear Assignment Problem (LAP) using the Jonker-Volgenant algorithm \cite{crouse2016implementing}.

\subsection{Scan Registration}\label{sec:registration}
\begin{figure}[htp]
  \centering
    \includegraphics[width=\columnwidth]{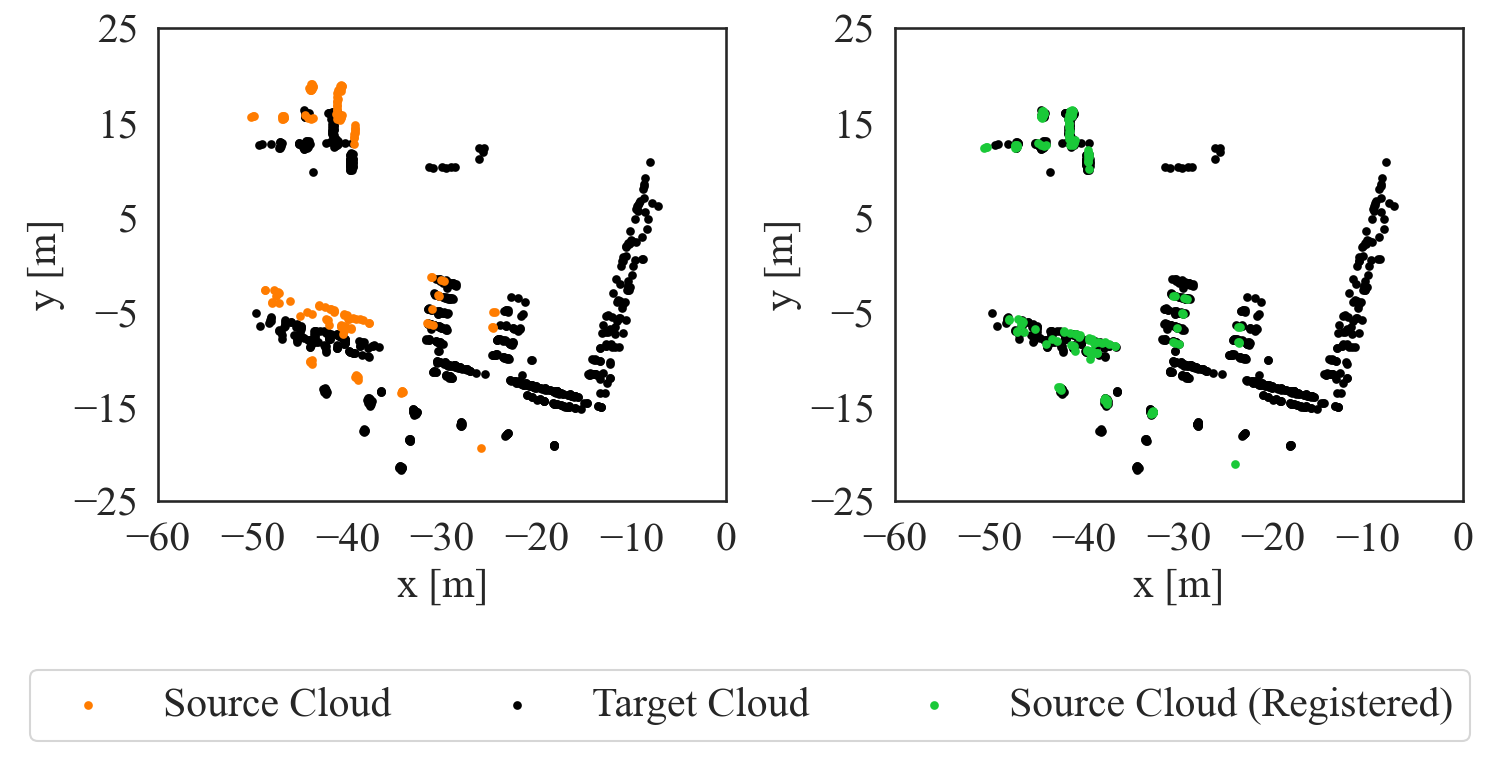}%
  \caption{\textbf{Graph matching-based ICP registration. %on the USMMA dataset.
  } The source cloud (orange) is roughly transformed into the local robot coordinate system using the transformation estimated from object graph matching. It is then aligned with the target cloud (black) using ICP with a sliding window of size 3. The registered source cloud is shown in green.}
  \vspace{-5mm}
\label{fig:matching}%
\end{figure}

When more than three matched object pairs are obtained from the graph matching process in Sec.~\ref{sec:graph}, the coordinate transformation $ \mathbf{T}^\beta_\alpha $ from $ \mathcal{V}\beta $ to $ \mathcal{V}\alpha $ is computed using affine transformation estimation. To ensure robustness, we apply RANSAC and accept the transformation only if it has more than four inliers. 
With the transformation $ \mathbf{T}^{\beta}_{\alpha} $ estimated from object maps as geometric priors, we apply ICP registration in a manner similar to its application in single-robot SLAM \cite{wang2022virtual}.
We use a sliding window for the target cloud to improve the registration results. Fig.~\ref{fig:matching} illustrates the source and target clouds before and after ICP registration, demonstrating that the initial guess provided by object graph matching is sufficient for successful ICP registration.
\subsection{Group-wise Consistent Measurement Set Maximization}
\begin{figure}[htp]
  \centering
        \includegraphics[width=\columnwidth]{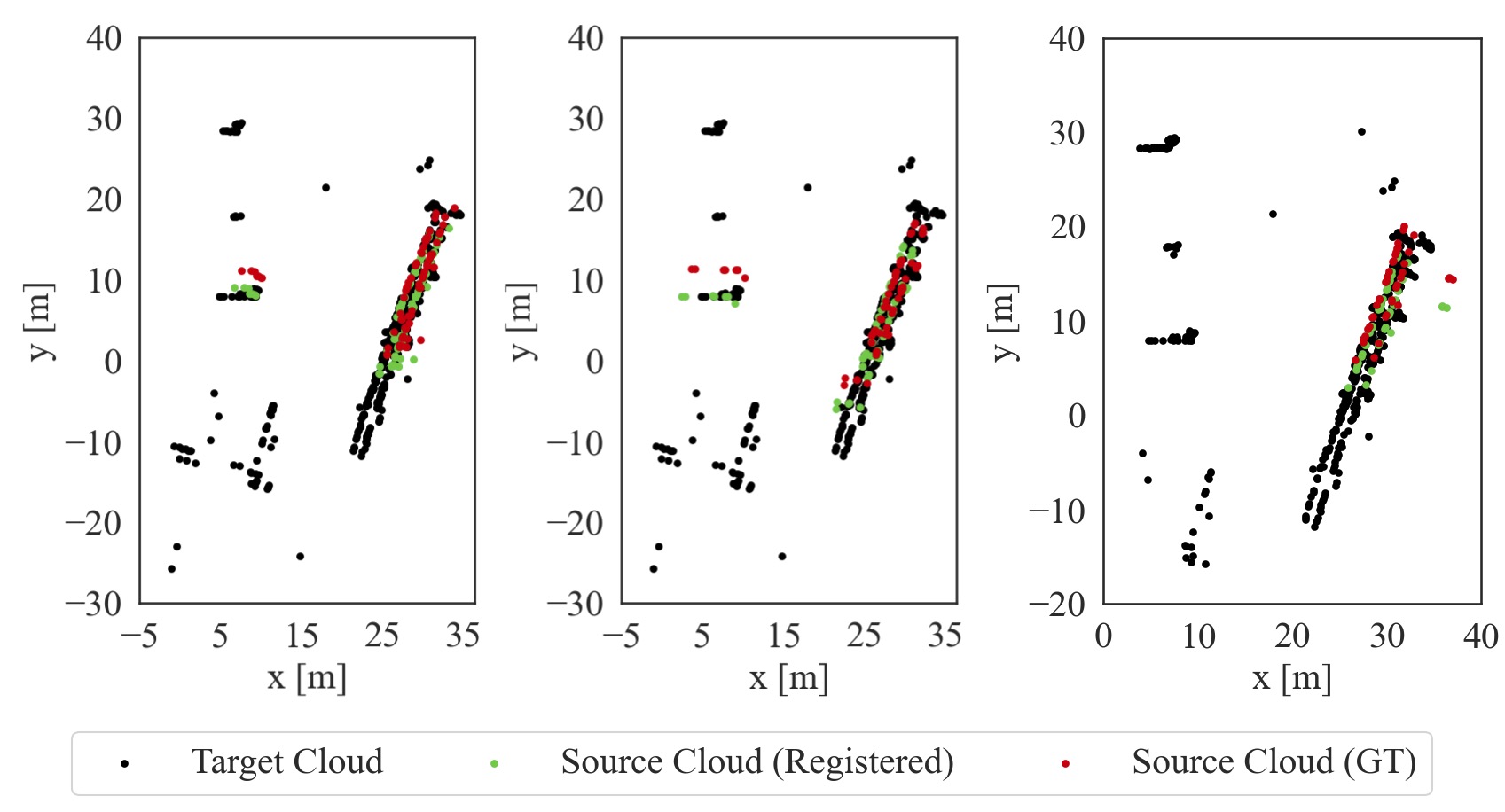}
  \caption{\textbf{Motivation for use of GCM.} A series of three inter-robot loop closures, in the same region of the environment (from our USMMA dataset depicted in Fig. \ref{fig:usmma_real}), are impacted by similar point cloud registration errors. Each erroneous registration result is shown in green, with ground truth shown in red.} \label{fig:error-icp}
\end{figure}

A challenging aspect of scan registration is the fact that separate pairs of scans located in close proximity to one another can give rise to similar registration errors,
%shared errors among nearby loop closures, 
as illustrated in Figure~\ref{fig:error-icp}. For sonar scan registration, the overlap percentage between the source and target point clouds is an intuitive indicator of quality. 
However, as we will discuss in Section~\ref{sec:icp}, even a high overlap ratio does not always guarantee minimal registration error.
While this is acceptable when using a global PCM, it can lead to errors when only incremental PCM is applied.
To address these challenges, we propose a modified approach that we denote Group-wise Consistent Set Measurement Maximization (GCM), inspired by PCM.

\begin{figure}[htp]
  \centering
    \subfigure[PCM]{
    \centering
        \includegraphics[width=.25\columnwidth]{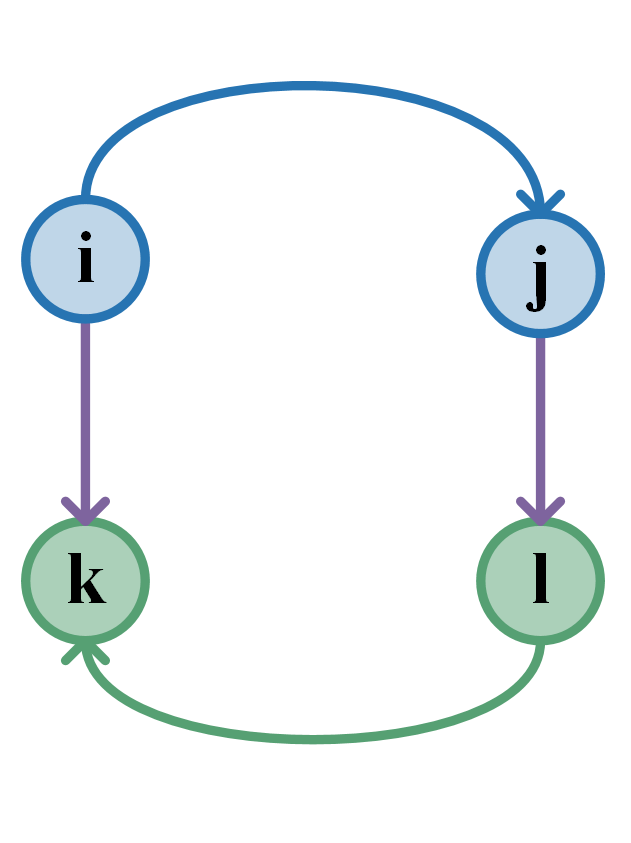}\label{fig:pcm}
    } \hspace{15pt}
    \subfigure[GCM]{
    \centering
        \includegraphics[width=.48\columnwidth]{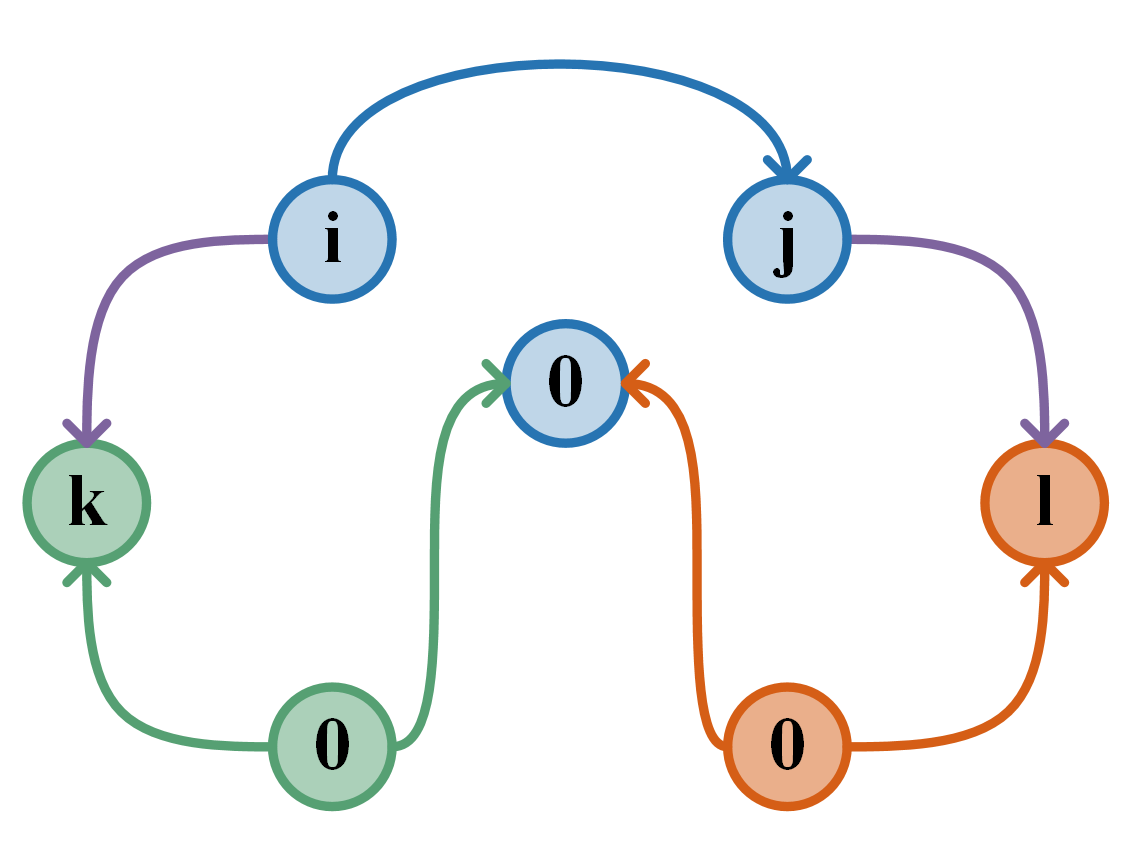}\label{fig:gcm}
    } 
  \caption{\textbf{Illustration comparing \textbf{PCM} and the proposed \textbf{GCM}}. Circles represent pose estimations, arrows indicate measurements, and purple arrows highlight inter-robot loop closure measurements.} \label{fig:pcm-gcm}\vspace{-5mm}
\end{figure}

In PCM \cite{mangelson2018pairwise}, the consistency of observations between a pair of robots is considered, as illustrated in Fig.~\ref{fig:pcm}.
For two inter-robot loop closure observations, $\mathbf{z}^{\beta_i}_{\alpha_k}$ and $\mathbf{z}^{\beta_j}_{\alpha_l}$, involving robot $\alpha$ and robot $\beta$, the method evaluates their consistency using:
\begin{equation}
    C(\mathbf{z}^{\beta_i}_{\alpha_k}, \mathbf{z}^{\beta_j}_{\alpha_l}) = \norm{(\mathbf{z}^{\beta_i}_{\alpha_k})^{-1} \cdot \hat{\mathbf{x}}^{\beta_i}_{\beta_j} \cdot \mathbf{z}^{\beta_j}_{\alpha_l} \cdot \hat{\mathbf{x}}^{\alpha_l}_{\alpha_k}}.
\end{equation}
% start here
Where $\hat{\mathbf{x}}^{\beta_i}_{\beta_j} $ and $ \hat{\mathbf{x}}^{\alpha_l}_{\alpha_k} $ are the relative state estimations marginalized from the factor graph optimization of local SLAM.

In the proposed GCM, we consider the consistency of observations among a group of robots (Fig.~\ref{fig:gcm}).
For two inter robot loop closure observations, $\mathbf{z}_{\alpha_k}^{\beta_i}$ between local robot $\alpha$ and local robot $\beta$ and $\mathbf{z}_{\alpha_l}^{\gamma_j}$ between robot $\alpha$ and robot $\gamma$, we perform:
\begin{equation}
    C(\mathbf{z}^{\beta_i}_{\alpha_k}, \mathbf{z}^{\gamma_j}_{\alpha_l}) = \norm{(\mathbf{z}^{\beta_i}_{\alpha_k})^{-1} \cdot \hat{\mathbf x}^{\beta_i}_{\gamma_j} \cdot \mathbf z^{\gamma_j}_{\alpha_l} \cdot \hat{\mathbf x}^{\alpha_l}_{\alpha_k} }.
\end{equation}
Robot $\alpha$ is considered the local robot, so $\hat{\mathbf{x}}^{\alpha_l}_{\alpha_k}$ can be obtained by marginalizing the factor graph optimization of local SLAM. We further expand $\hat{\mathbf{x}}^{\gamma_l}_{\beta_k}$:
\begin{align}
    \hat{\mathbf x}^{\beta_i}_{\gamma_j} & = (\hat{\mathbf x}^{\alpha_0}_{\beta_i})^{-1}  \cdot \hat{\mathbf x}^{\alpha_0}_{\gamma_l},\\
    & = (\hat{\mathbf x}^{\beta_0}_{\beta_i})^{-1} \cdot \hat{\mathbf x }^{\beta_0}_{\alpha_0} \cdot 
    (\hat{\mathbf x}^{\gamma_0}_{\alpha_0})^{-1}\cdot \hat{\mathbf x}^{\gamma_0}_{\gamma_j}.
\end{align}
The relative state estimations $\hat{\mathbf{x}}^{\gamma_0}_{\gamma_l}$ and $\hat{\mathbf{x}}^{\beta_0}_{\beta_k}$ can be marginalized from the local factor graph optimization of neighboring robots, respectively. The relative state estimations  $\hat{x}^{\beta_0}_{\alpha_0}$ and $\hat{x}^{\gamma_0}_{\alpha_0}$ are the optimized map coordinate transformations obtained from historically accepted loop closures.

\subsection{Inter-robot PGO}\label{sec:inter-PGO}
Let $\mathcal{R} = \{\alpha, \beta, \gamma, ... \}$ represent the set of robots in the team. Similar to the formulation in Eq.~\eqref{eqn:local}, the inter-robot pose graph optimization for the entire robot team can be expressed as:
\begin{align}
    \mathcal{X}^*_{\mathcal{R}} &= \arg \max_{\mathcal{X}^*_{\mathcal{R}}} P(\mathcal{X}^*_{\mathcal{R}} \mid \mathcal {Z}_{\mathcal{R}}),\label{eqn:full}\\
    \mathcal{Z}_{\mathcal{R}} &= \mathcal{Z}_{\mathcal{R}}^{\text{local}} \cup \mathcal{Z}_{\mathcal{R}}^{\text{inter}}.
\end{align}
Here, $\mathcal{X}_{\mathcal{R}} = \{\mathcal{X}_{i} \mid i \in \mathcal{R}\}$ represents the states for all robots in the team. 
The set of observations for the entire team is denoted by:
\begin{equation}
    \mathcal{Z}_{\mathcal{R}} = \mathcal{Z}_{\mathcal{R}}^{\text{local}} \cup \mathcal{Z}_{\mathcal{R}}^{\text{inter}},
\end{equation}
where $\mathcal{Z}_{\mathcal{R}}^{\text{local}} = \{\mathcal{Z}_i \mid i \in \mathcal{R}\}$ represents the local observations from each robot,
and $\mathcal{Z}_{\mathcal{R}}^{\text{inter}} = \{\mathcal{Z}_{i,j} \mid i,j \in \mathcal{R}, i \neq j\}$ represents the inter-robot loop closures.

Although full PGO in Eq.~\eqref{eqn:full} can yield good results by leveraging all available local and global information, it involves solving a large-scale optimization problem. 
This process can require significant time to converge, particularly when the initial estimate lacks accuracy or when the number of robots in the team grows.

To address this challenge, we apply a two-step approach combining global and local PGO, beginning with the \textbf{local optimization}:
\begin{equation}
    \mathcal{X}^*_{\alpha}, \mathcal{X}^*_{\alpha, \mathcal R} = \arg \max_{\mathcal{X}_{\alpha}, \mathcal{X}_{\alpha, \mathcal R}} P(\mathcal{X}_{\alpha}, \mathcal {X}_{\alpha, \mathcal R} \mid \mathcal {Z}_{\alpha}, \mathcal {Z}_{\alpha, \mathcal{R}}).\label{eqn:step1}
\end{equation}
Here, $\mathcal{Z}_{\alpha, \mathcal{R}} = \bigcup\limits_{\substack{i \in \mathcal{R} \ i \neq \alpha}} \mathcal{Z}_{\alpha, i}$ represents the inter-robot loop closure observations between the local robot $\alpha$ and all its neighboring robots.
$\mathcal{X}_{\alpha, \mathcal{R}}$ denotes the states from the neighboring robots that share inter-robot loop closures with the local robot.

% %%%%%%%%%%%%%%%%%%%%%
Next, we proceed with the \textbf{global optimization} for neighboring robots involved in inter-robot data association.
Consider a neighboring robot $\beta$ that shares inter-robot loop closures $\mathcal{Z}_{\alpha, \beta} \subset \mathcal{Z}_{\alpha, \mathcal{R}}$ with the local robot $\alpha$. The states of $\beta$ can be optimized as follows:
\begin{equation}
    \mathcal{X}^*_{\beta} = \arg \max_{\mathcal{X}_{\beta}} P(\mathcal{X}_{\beta} \mid \hat{\mathcal{X}^o_{\beta}}, {\mathcal{X}}^*_{\alpha, \mathcal \beta}),
\end{equation}
where ${\mathcal{X}}^*_{\alpha, \beta} \subset \mathcal{X}^*_{\alpha, \mathcal{R}}$ represents the optimized subset of historical states of robot $\beta$, obtained from Eq.~\eqref{eqn:step1}, that are associated with robot $\alpha$, and $\hat{\mathcal{X}^o_{\beta}}$ represents the estimated relative states marginalized from the local SLAM of the neighboring robot $\beta$.

All initial guesses for both local and global PGO are obtained by applying map coordinate transformations based on object graph matching.
Inter-robot loop closures are added to the factor graph using a robust noise model to ensure accurate and reliable association. 
  By adopting this two-step PGO approach, we enable faster convergence of the factor graph optimization. This is achieved by updating only a subset of the information at each step, rather than recalculating the entire graph, which reduces computational overhead and accelerates the optimization process.

\subsection{Communication}
In this section, we summarize the communication overhead between the local robot and its neighboring robots. 
As depicted in Fig.~\ref{fig:pipeline}, at each sonar timestep, the object map is shared with the robot team as an objects message. Additionally, pose estimations are incrementally transmitted to neighboring robots.

To optimize the use of communication bandwidth, updates to the historical state estimations for neighboring robots are only performed when significant changes result from loop closures. 
%A similar strategy could be applied to the object map in future work to further improve bandwidth efficiency.
Once the graphs are matched, the sonar scans associated with the matched objects are requested from the neighboring robots. 
After the detection of inter-robot loop closures, this information is also shared with the robot neighbors.
 
\section{Experiments and Results}\label{sec:setup}
We evaluate DRACo-SLAM2 using both real-world and simulated data. The real-world data, shown in Fig.~\ref{fig:real}, was collected using our customized BlueROV2-Heavy (illustrated in Fig.~\ref{fig:pipeline}) at the U.S. Merchant Marine Academy (USMMA), King's Point, NY. 
Details of the BlueROV2 configuration can be found in our previous work \cite{mcconnell2022draco}. 
For this study, we utilize the robot's horizontally-oriented Oculus M750d sonar, along with a Rowe SeaPilot DVL and a VectorNav VN100 MEMS IMU.
Two fully simulated datasets, the USMMA dataset and the airplane dataset, were generated using HoloOcean \cite{HoloOcean}. %the UUV simulator~\cite{manhaes2016uuv} within the Gazebo environment~\cite{koenig2004design}. 
The USMMA dataset replicates key features of the environment where the real USMMA dataset was collected, including floating docks and repeating circular pier pilings. 
The airplane dataset is intended to simulate the site of an airplane wreck located on the seafloor (this dataset is shown in Fig. \ref{fig:pipeline}). In all datasets, both real and synthetic, all robots operate at the same, fixed depth (close to the surface in the USMMA datasets, and close to the seafloor in the airplane dataset). 
To simulate the simultaneous operation of multiple robots, we recorded several dataset sessions and replayed them concurrently on a single computer (for both our real and synthetic datasets), with simulated communications between robots.
All experiments were conducted on a single computer equipped with an Intel Xeon E-2276M CPU, 31.1 GB of memory, and running Ubuntu  20.04 with ROS Noetic. %\textcolor{red}{version?}.

\begin{figure}[ht]
  \centering
  \subfigure[Result on the airplane dataset.]{
      \includegraphics[width=.98\columnwidth]{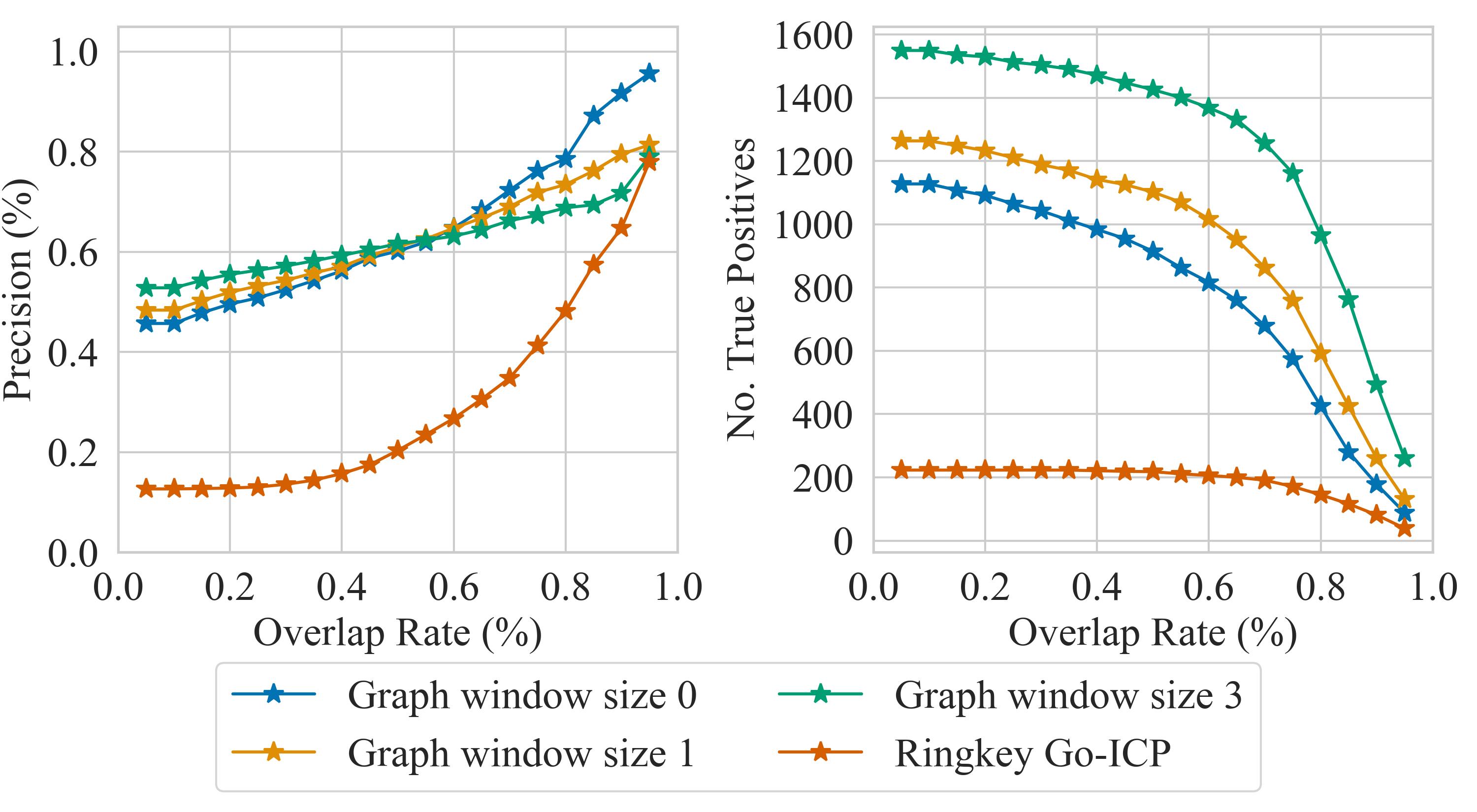}
      \label{fig:plane_icp}
  }

  \subfigure[Result on the USMMA dataset.]{
      \includegraphics[width=.98\columnwidth]{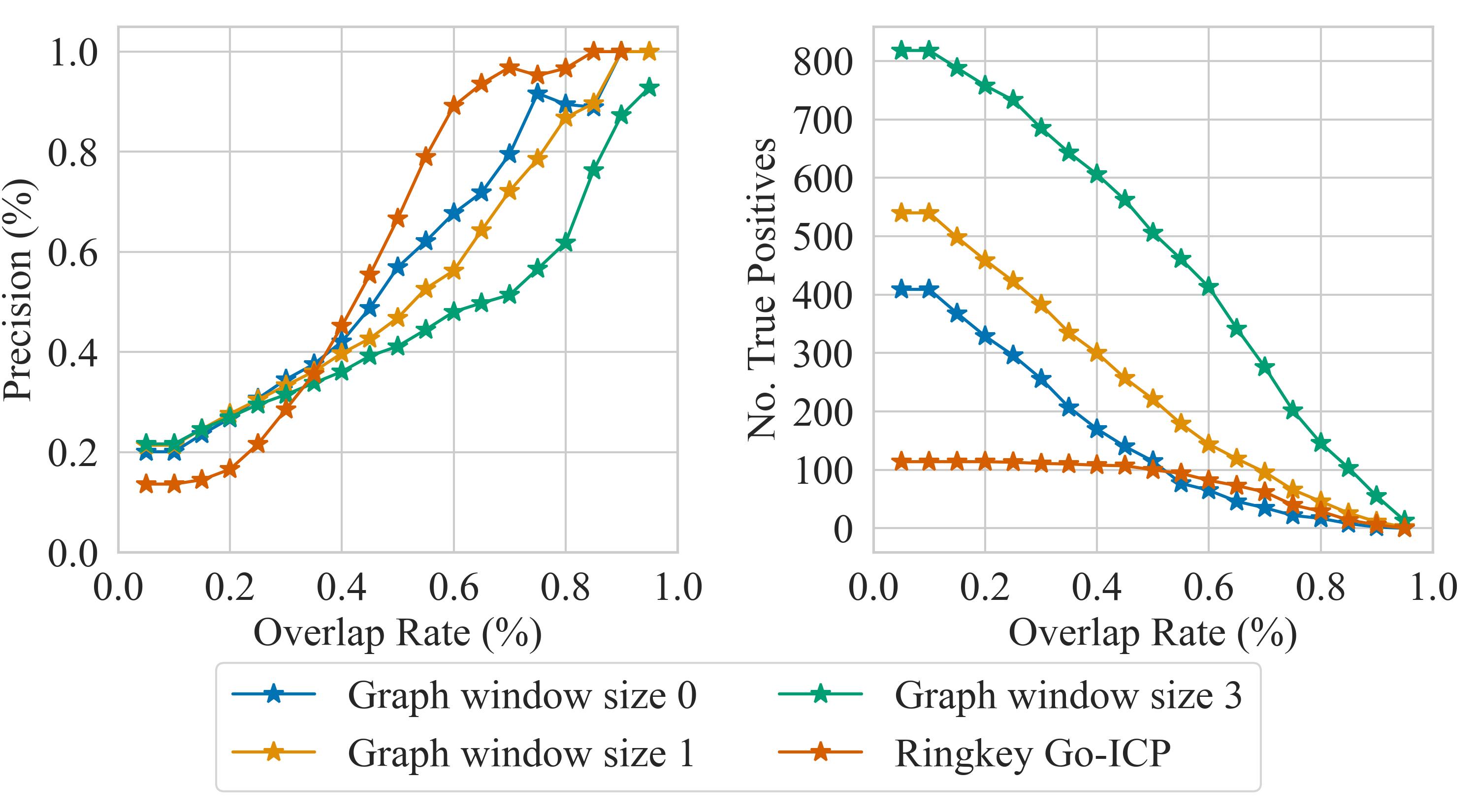}
      \label{fig:usmma_icp}
  }
  \caption{\textbf{Performance} of inter-robot loop closure detection and registration on both of our 3-robot fully simulated datasets. Precision (left) and the number of true positive loop closures detected (right) plotted against the overlap ratio parameter $r_{overlap}$, evaluated using various methods.}\label{fig:exp_icp} \vspace{-4mm}
\end{figure}

\subsection{Performance of Inter-Robot Loop Closure Detection}\label{sec:icp}
In this section, we evaluate the performance of our object-graph-matching-based inter-robot data association technique. Fig.~\ref{fig:exp_icp} shows the precision (left) and the number of correct loop closures detected (right) as a function of the minimum acceptable overlap ratio on our two fully simulated datasets.

The overlap ratio, $r_{\text{overlap}} = \frac{n_{\text{overlap}}}{n_{\text{total}}}$, measures the proportion of points in the target cloud ($n_{\text{overlap}}$) that overlap with the source cloud, relative to the total number of points in the target cloud ($n_{\text{total}}$).
A loop closure is considered a true positive if the estimated translation error is within 1.5 m and the angular error is within 15\degree ~of the ground truth.
Precision is defined as the ratio of true positives to the total detected loop closures.

We tested the proposed algorithm using different sliding window sizes. A window size of 0 indicates no sliding window, while sizes 1 and 3 correspond to the inclusion of 3 and 7 nearby frames for registration, respectively.
Across both datasets, the method with a sliding window of size 3 achieved the highest number of true positive loop closures. 
However, its precision decreased when the minimum overlap ratio threshold was relatively low, which is expected due to the sliding window strategy.
To balance these factors, we set the minimum acceptable overlap ratio threshold to $\epsilon_{\text{overlap}} = 0.9$ for all experiments and accepted a loop closure only if $r_{\text{overlap}} > \epsilon_{\text{overlap}}$.

\begin{table}[htp]
\caption{\textbf{Runtime} for sonar scene descriptor-based inter-robot loop closure and global ICP registration in DRACo-SLAM \cite{mcconnell2022draco} (referred to as DRACo1), and graph-based inter-robot loop closure detection and ICP registration in the proposed DRACo-SLAM2 (referred to as DRACo2) on both of our 3-robot fully simulated datasets.}\vspace{-3mm}
\renewcommand{\arraystretch}{1.2}
\label{tab:icp_runtime}
\begin{center}
\begin{tabular}{cccc}
\Xhline{2\arrayrulewidth}
\multicolumn{2}{c}{\multirow{2}{*}{\textbf{Algorithm}}} & \multicolumn{2}{c}{\textbf{Runtime (ms)}}\\
 & & \textbf{Mean} & \textbf{Max}\\
\Xhline{2\arrayrulewidth}
\multirow{2}{*}{DRACo1} & Ring Key Matching & 0.12 & 2.05 \\
& Go-ICP Registration & 361.83 & 2556.38 \\
\Xhline{1\arrayrulewidth}
\multirow{2}{*}{DRACo2} & Graph Construction \& Matching & 5.64 & 32.76 \\
& ICP Registration (Window Size 3) & 12.69 & 358.45 \\
\Xhline{2\arrayrulewidth}
\end{tabular}
\end{center}
\end{table}
\vspace{-3mm}
We also summarize the runtime of our proposed DRACo-SLAM2 method (referred to as DRACo2) and its predecessor method, DRACo-SLAM (referred to as DRACo1) in Tab.~\ref{tab:icp_runtime}. 
As shown in Tab.~\ref{tab:icp_runtime}, the ICP registration module in DRACo2 is, on average, 20 times faster than that in DRACo1 because the time-consuming global registration step is not required in our approach.
This enables us to operate at a much higher frequency and detect significantly more loop closures compared to the original method.
As shown in Tab.~\ref{tab:timestep-results}, our total loop closure detection module achieves a runtime per timestep that is 10 times faster than DRACo1, enabling more time-efficient performance.

\begin{table}[htp]
\caption{\textbf{Mean runtime and mean number of inter-robot loop closure algorithm executions per time step} for DRACo1 and DRACo2 on both of our 3-robot fully simulated datasets.}\vspace{-3mm}
\renewcommand{\arraystretch}{1.2}
\label{tab:timestep-results}
\begin{center}
\begin{tabular}{cccc}
\Xhline{2\arrayrulewidth}
\multicolumn{2}{c}{\textbf{Algorithm}} & \textbf{Mean (s)} & \textbf{Mean No.} \\
\Xhline{2\arrayrulewidth}
\multirow{2}{*}{DRACo1} & Ring Key Matching & $<$ 0.01 & 1 \\
& Go-ICP Registration & 10.11 & 28 \\
\Xhline{1\arrayrulewidth}
\multirow{2}{*}{DRACo2} & Graph Construction \& Matching & $<$ 0.01 & 1 \\
& ICP Registration (Window Size 3) & 1.55 & 122  \\
\Xhline{2\arrayrulewidth}
\end{tabular}
\end{center}
\end{table}
\vspace{-4mm}
\subsection{Performance of Inter-Robot PGO}
\begin{figure}[htp]
    \centering
  % \subfigure{
  %   \includegraphics[width=.41\columnwidth]{fig/plane_1_self.png}
  %     \label{fig:plane_traj}
  % }
  % \subfigure{
  % \includegraphics[width=.53\columnwidth]{fig/usmma_3_self.png}
  %     \label{fig:usmma_traj}
  % }
    \subfigure{
    \includegraphics[width=.6\columnwidth]{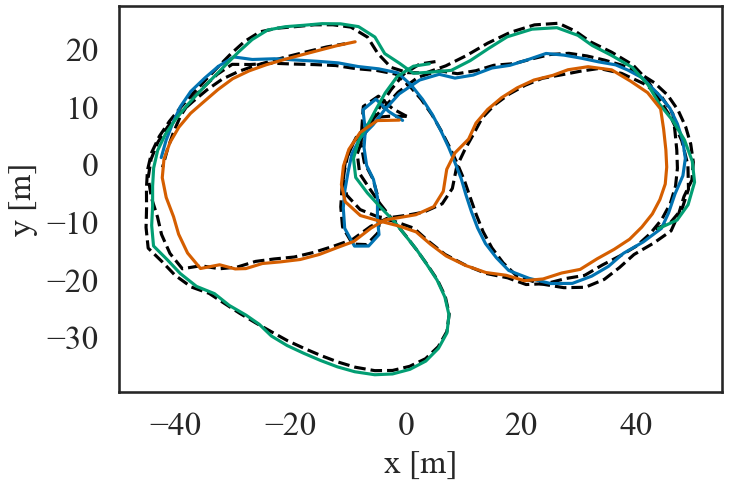}\label{fig:plane_traj}
  }\vspace{-1mm}
  \subfigure{
  \includegraphics[width=.595\columnwidth]{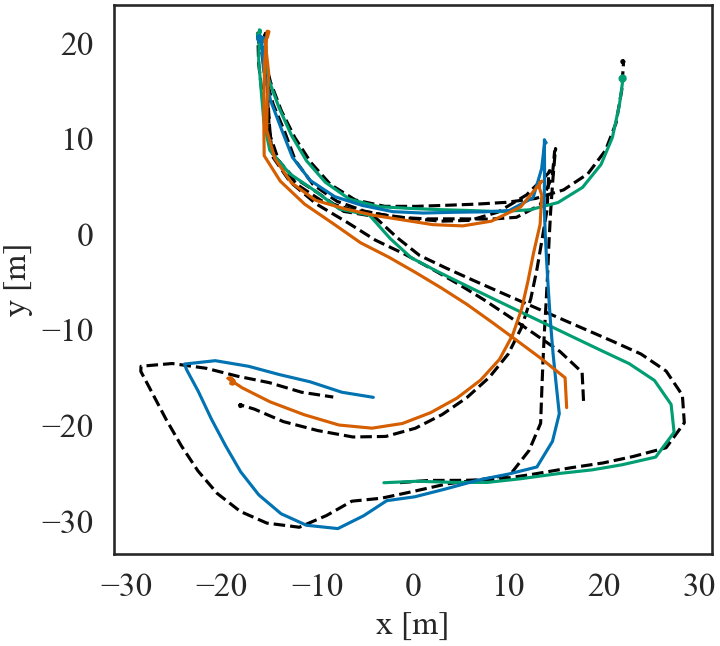}\label{fig:usmma_traj}
  }\vspace{-2mm}
  \subfigure{
  \centering
  \includegraphics[width=.98\columnwidth]{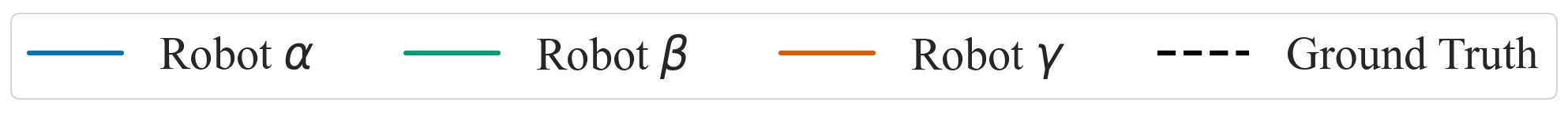}
  }\vspace{-1mm}
  \caption{\textbf{Optimized trajectories} of different local robots using the proposed DRACo-SLAM2 on the fully simulated 3-robot airplane (top) and USMMA (bottom) datasets.}\label{fig:trajectory}
\end{figure}

We also evaluate the performance of inter-robot PGO using both PCM and GCM on our two 3-robot fully simulated datasets.
A robust Cauchy noise model \cite{lee2013robust} from GTSAM \cite{gtsam} is employed for all methods to ensure robust performance.
Fig.~\ref{fig:trajectory} shows the trajectories optimized locally from robot $\alpha$ on the airplane dataset (left) and the USMMA dataset (right). 
We align the estimated trajectory with the ground truth using EVO \footnote{https://github.com/MichaelGrupp/evo}, treating the robot trajectories from different robots as a single trajectory with different timestamps. 
The trajectories are generally smooth, although drifts caused by local SLAM are observed in the results for the USMMA dataset.

\begin{table}[htp]
\caption{\textbf{Absolute Trajectory Error (ATE)} in meters for the proposed method compared to the full Pose Graph Optimization (PGO) method on both of our 3-robot fully simulated datasets.}\vspace{-3mm}
\renewcommand{\arraystretch}{1.2}
\label{tab:ate-results}
\begin{center}
\begin{tabular}{lcccccccc}
\Xhline{2\arrayrulewidth}
\multicolumn{2}{c}{\multirow{2}{*}{\textbf{Algorithm}}} & \multicolumn{3}{c}{\textbf{USMMA}}  & \multicolumn{3}{c}{\textbf{Airplane}}\\ 
 &  & \textbf{$\alpha$} & \textbf{$\beta$} & \textbf{$\gamma$} & \textbf{$\alpha$} & \textbf{$\beta$} & \textbf{$\gamma$}\\ 
\Xhline{2\arrayrulewidth}
\multirow{2}{*}{DRACo} & PCM & 1.58 & 2.43 & 1.88 & 1.32 & 1.36 & 1.46\\ 
 & GCM & 1.43 & \textbf{1.20} & \textbf{1.23} & 1.33 & \textbf{0.95} & \textbf{1.31}\\ 
\Xhline{1\arrayrulewidth}
\multirow{2}{*}{Full PGO} & PCM & 1.27 & 1.37 & 1.52 & \textbf{1.28} & 1.91 & 1.63\\ 
 & GCM & \textbf{1.24} & 1.41 & 1.56 & 1.31 & 1.91 & 1.65\\ 
\Xhline{2\arrayrulewidth}
\end{tabular}
\end{center}
\end{table}
\vspace{-3mm}
Additionally, we perform a quantitative analysis. 
Tab.~\ref{tab:ate-results} presents the Root Mean Square Error (RMSE) of the Absolute Trajectory Error (ATE) for our proposed two-step PGO method (referred to as DRACo) compared to the widely used full PGO with PCM and the proposed GCM. 
The ATE is also calculated using EVO.
For each robot, the estimated and ground truth trajectories are concatenated head-to-tail, starting with the trajectory of the local robot, followed by the trajectories of neighboring robots in alphabetical order.
Since EVO is designed for 6DoF SLAM, we adapt it for the 3DoF condition by setting $z = 0$, $\phi = 0$, and $\theta = 0$. 

As shown in Tab.~\ref{tab:ate-results}, our default configuration, DRACo two-step PGO with GCM, achieves the best accuracy in most cases. 
This is attributed to the introduction of GCM, which efficiently reduces the influence of loop closures with similar registration errors, an issue that remains inevitable even with a robust noise model. 
Furthermore, the adoption of the two-step PGO prevents drifts from neighboring robots from affecting the results of the local robot. 
However, as a trade-off, the improved accuracy from neighboring robots has only a limited influence on the state estimation of the local robot.

\vspace{-4mm}
\begin{table}[htp]
\caption{\textbf{Sizes of perception messages} for DRACo1 and DRACo2 on the USMMA real-world dataset (with simulated 3-robot comms.).}\vspace{-3mm}
\renewcommand{\arraystretch}{1.2}
\label{tab:comm-results}
\begin{center}
\begin{tabular}{lccc}
\Xhline{2\arrayrulewidth}
\textbf{Algorithm} & \textbf{Message Type} & \textbf{Mean KBits} & \textbf{Max KBits}\\
\Xhline{2\arrayrulewidth}
\multirow{2}{*}{DRACo1} & Ring key Descriptor & 0.13 & 0.13 \\
& Point Cloud-float32 & 9.67 & 27.90 \\
\Xhline{1\arrayrulewidth}
\multirow{2}{*}{DRACo2} & Object Map & 1.35 & 2.25 \\
& Point Cloud-float32 & 9.69 & 27.90 \\
\Xhline{2\arrayrulewidth}
\end{tabular}
\end{center}
\end{table}
\vspace{-4mm}
\subsection{Performance on the Real-world Dataset}
The proposed algorithm is next evaluated on our real-world dataset from USMMA with simulated 3-robot communications. 
Fig.~\ref{fig:real} illustrates the optimized trajectories, with results from robots $\alpha$, $\beta$, and $\gamma$ marked in blue, green, and orange, respectively. All detected inter-robot loop closures are indicated in purple for clarity.
Since ground truth data is unavailable for this real-world dataset, we align the constructed point-cloud map with satellite imagery for a more intuitive and visually interpretable representation of the results. 

Tab.~\ref{tab:comm-results} compares the sizes of perception messages transmitted by the full version of DRACo-SLAM (DRACo1) and DRACo-SLAM2 (DRACo2) on the USMMA real-world dataset. For DRACo1, the Ring Key Descriptor messages used for initial loop closure candidate detection are compact, with both mean and maximum sizes under $1$ KBits. 
Object Map messages introduced by DRACo2 are larger but remain well below the 62.5 kbps bandwidth limit of HS underwater acoustic modems operating at a range of $300$m. 
The communication bandwidth required for point-cloud transmission is similar for both methods.

\vspace{-4mm}
\section{Conclusion}\label{sec:conclusion}
In this paper, we propose DRACo-SLAM2, a distributed multi-robot SLAM framework designed for sonar-equipped underwater robot teams using object graph matching. By clustering the point-cloud map into an object map, exchanging the object maps, and comparing them through graph matching, DRACo-SLAM2 achieves a tenfold speed improvement over our previous work, DRACo-SLAM.
Additionally, we introduce the Group-wise Consistent Measurement Set Maximization (GCM) technique to enhance outlier rejection, and address the challenges of similar registration errors across multiple sonar image pairs encountered in underwater mapping scenarios.
For future work, we aim to optimize the data updating strategy for object map exchanges, and integrate the algorithm with an autonomous exploration algorithm to enable real-world underwater exploration with robot teams.
\vspace{-2mm}



% \section*{ACKNOWLEDGMENT}

% The preferred spelling of the word ÒacknowledgmentÓ in America is without an ÒeÓ after the ÒgÓ. Avoid the stilted expression, ÒOne of us (R. B. G.) thanks . . .Ó  Instead, try ÒR. B. G. thanksÓ. Put sponsor acknowledgments in the unnumbered footnote on the first page.

%%%%%%%%%%%%%%%%%%%%%%%%%%%%%%%%%%%%%%%%%%%%%%%%%%%%%%%%%%%%%%%%%%%%%%%%%%%%%%%%

\bibliographystyle{IEEEtran}
\bibliography{bib}

\end{document}